# Analysis of PDE-based binarization model for degraded document images

**Uche A. Nnolim***

*Department of Electronic Engineering, Faculty of Engineering, University of Nigeria, Nsukka, Enugu, Nigeria*

uche.nnolim@unn.edu.ng

**Abstract:** *This report presents the results of a PDE-based binarization model for degraded document images. The model utilizes an edge and binary source term in its formulation. Results indicate effectiveness for document images with bleed-through and faded text and stains to a lesser extent.*

## 1. Introduction

Text binarization is essential in document image processing systems and is crucial for optical character recognition (OCR) systems. Recently, deep learning schemes have been steadily applied to the field of document image processing. However, the computational cost is high in data, time and computing resources for effective results. Also, PDE-based models have been steadily growing in the literature. This is due to their relatively low complexity, simplicity and effectiveness compared to their complexity [1]. However, most models still have problems in effectively binarizing highly degraded document images and only recently have researchers been more focused on addressing these issues [1] [2] [3].

The outline of the work is as follows; section 2 presents a background and brief overview of PDE- and deep learning-based binarization methods for document images. Section 3 presents the proposed scheme, while section deals with results and comparisons. The final section presents the conclusions.

## 2. Background and motivation

The Deep learning based methods for document image binarization include the selectional auto-encoder method by Calvo-Zaragoza and Gallego [4], global-local UNets by Huang et al [5], integrated multiple pre-trained U-Net modules by Kang et al [6], DeepOtsu-based iterative deep learning by He and Schomaker [7], combined local features with support vector machine (SVM) by Xiong et al [8] and primal-dual network (PDNet) by Ayyalasomayajula et al [9], cascaded generators of conditional generative adversarial networks by Zhao et al [10], Fully convolutional networks by Long et al [11], Pastor-Pellicer et al [12] and Ronneberger et al [13].

The PDE methods include models by Cheriet [14], Nwogu et al [14], Mahani et al [14], Kumar et al [15], Bella et al [16], Guemri and Drira [17], Wang et al [18], Wang and He [19], Jacobs and Momoniat [20] [21], Rivest-Henault et al [22], Drira and Yagoubi [14], Chen et al [23], Huang et al [24], Guo et al [14] [25], Guo and He [26], Zhang et al [27], Feng [28], Nnolim [1] [2] [3] and Du and He [29].

In most of the PDE-based works, the standard approaches involve including a binarization source term and a diffusion term. However, the scheme by Du and He contains no source term but rather diffusion and contrast enhancement terms. Also, the proposed scheme only possesses an edge term and a source term [1]. Several of the models were analyzed before selecting the version Zhang et al [27], as the model that could be further improved and was more versatile due to the structure tensor component based on the hessian matrix [27]. Also, several diffusivity functions from the literature were analyzed prior to selecting the best for degraded document images, which was further improved upon.

## 3. Proposed model

Given that the model needed to satisfy the need for simplicity, intuitiveness and easy to tune or control, simplifications using local monotonic (LOMO) diffusion by Acton was performed to reduce computational burden [1]. However, the signed edge detector lost its sensitivity due to the signum operator, which had to be discarded to realize the best outcomes [1]. Also, the source term must be limited to the range of -1 to +1 in order for the model to satisfy boundary conditions. The proposed (integer order) PDE model was given [1] as;

$$\frac{\partial u}{\partial t} = c_s\left[\left(\frac{1}{a}\right)\tan^{-1}(u)\right] + c_e\left[1 - \frac{p}{1+q*\left(\frac{h*min_{\Omega}(h)}{k}\right)^2}\right] \quad (1)$$

Where $;0 \leq a < 1$ and $p > 0; q > 0$ are parameters which control the binarization term and the diffusivity function in the edge term respectively [1]. Detailed analysis of the functions, its parameters and others from the literature were performed to gauge performance and aid in the selection criteria for utilization in the proposed model [1]. Furthermore, a fractional version of the PDE was also derived and its numerical implementation was developed [1]. The nature of the model with default setting of the binary source coefficient, $c_s = 1$, allows direct control of results by mainly the edge term, whose contribution is regulated by the parameter, $c_e$. The model performed well for degradations whose profiles had weak edges or smooth profiles [1]. However, problems arose when stains had similar edge profiles similar to desired foreground text.

## 4. Experiments and comparisons

Extensive experiments were performed to test the versatility of the proposed model with regard to parameter tuning [1]. Then performance comparison was done with other existing PDE-based approaches, which included methods by Wang and He (WH) [19], Rivest-Hénault et al (RMC) [22], Jacobs and Momoniat (JM) [20], Wang et al (WYH) [18], FENG [28], Guo et al (GHZ) [14], Zhang et al (ZHG) [27],

Guo et al (GHW) [25]. The first figure deals with the analysis of some colour degraded document images, which require conversion to HSI domain for effective binarization compared to standard grayscale conversion [1]. The second figure presents results using dynamic threshold, depicting the threshold curves for sample document images.

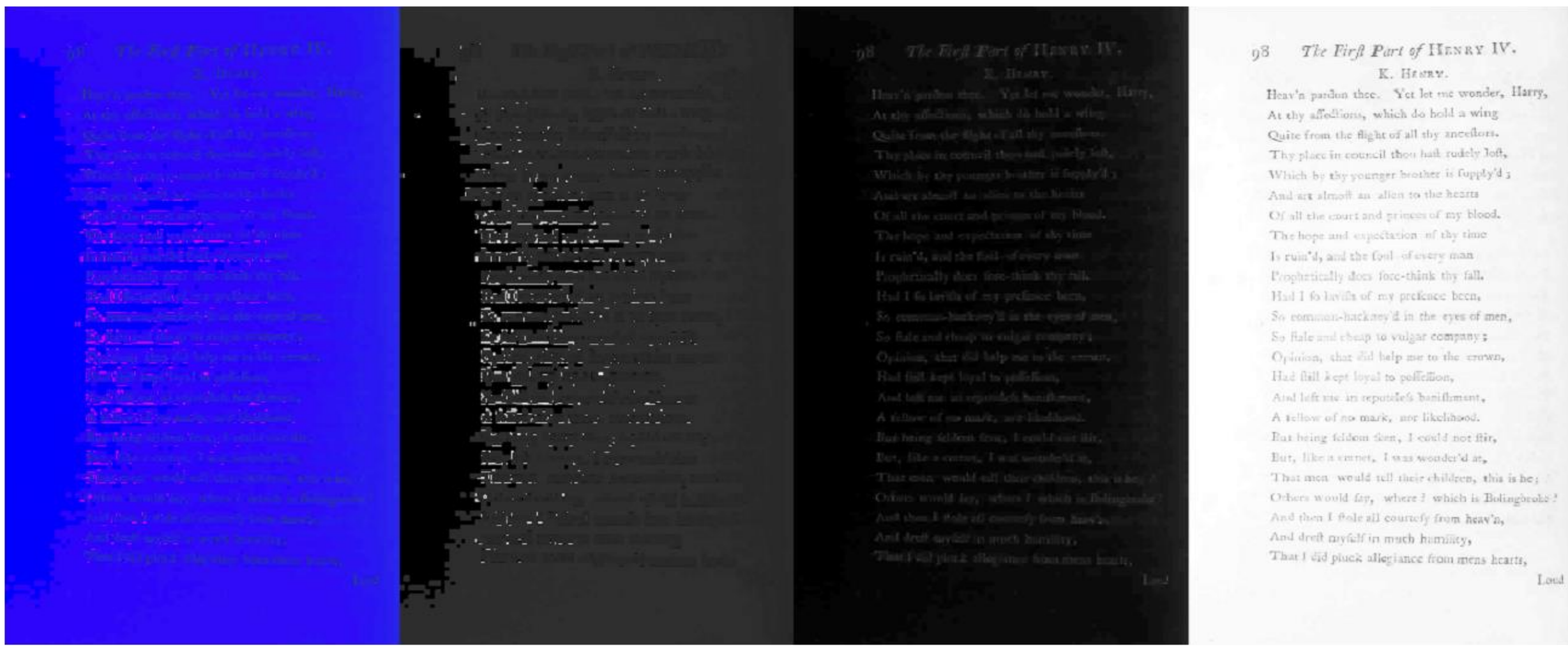

(a) (b) (c) (d)

Fig. 1. HSI image (a) hue (b) saturation (c) intensity channel analysis of image 17 from DIBCO2017 series

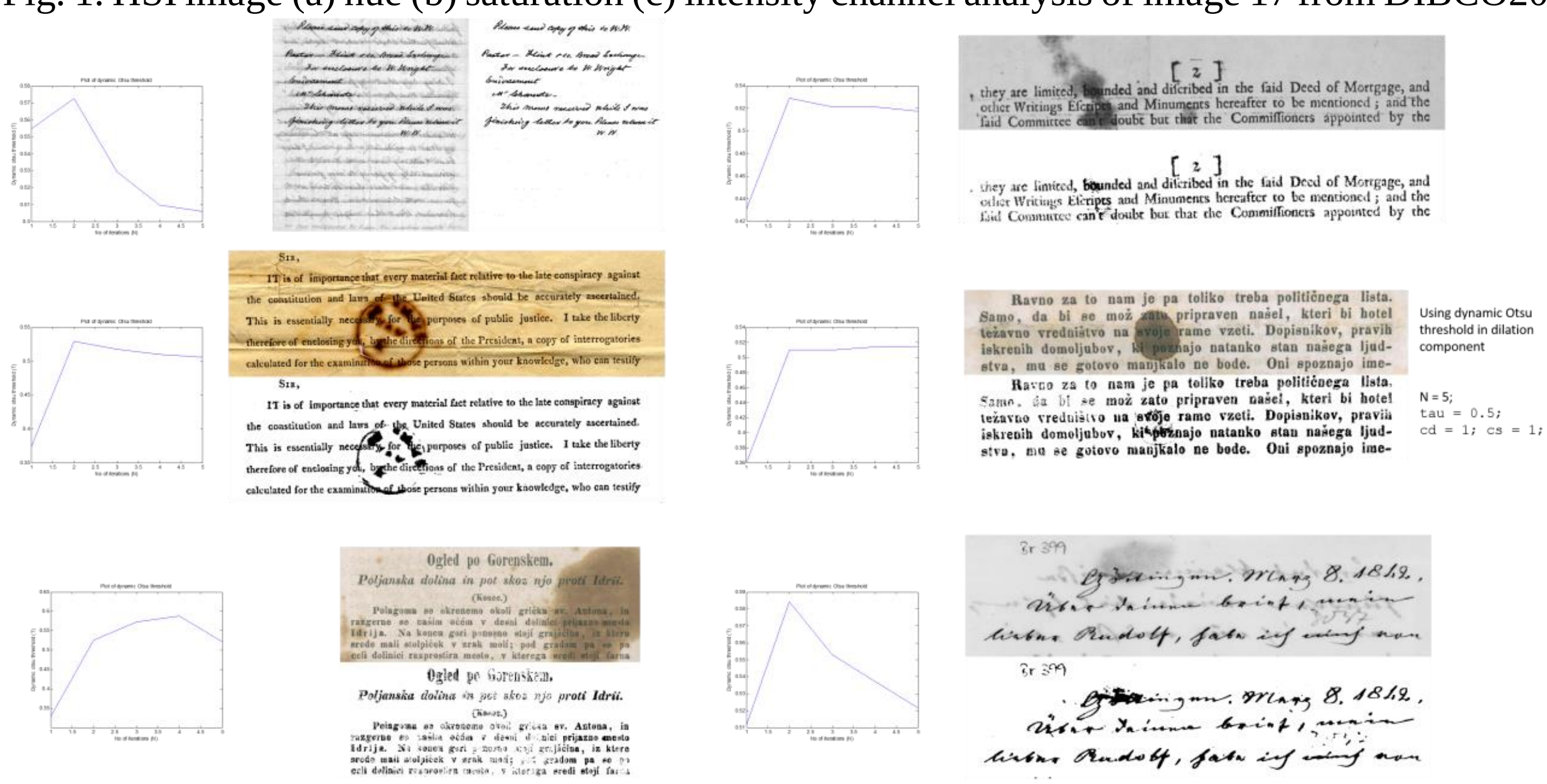

Fig. 2. Curves for PA using dynamic threshold for various document images with different degradations

The visual results of PA are shown in Fig. 3 to Fig. 6.

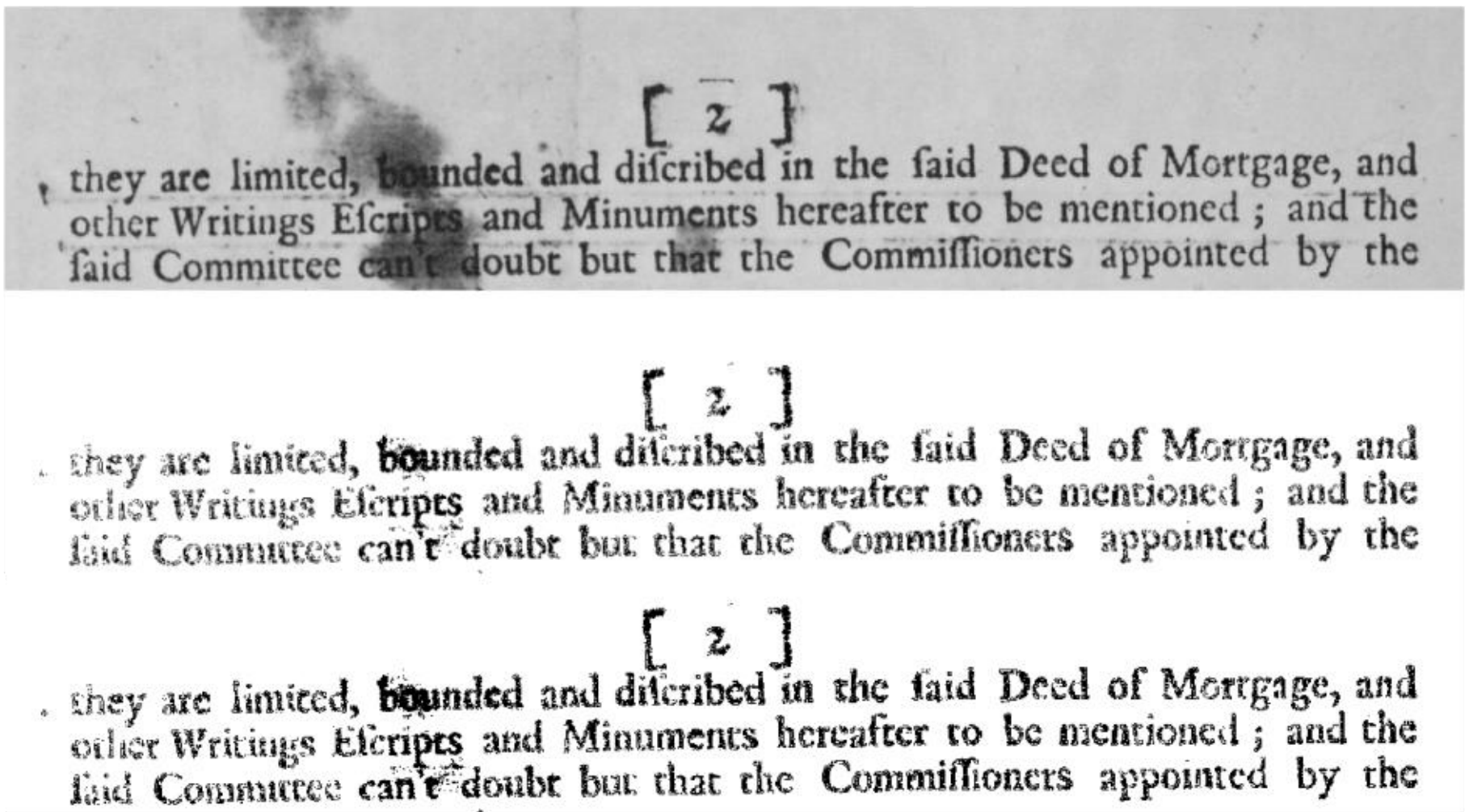

N = 10;
tau = 0.25;
cd = 0.95; cs = 1;

N = 5;
tau = 0.5;
cd = 1; cs = 1;

Fig. 3. Results of PA using various parameter settings for degraded document image with small stains

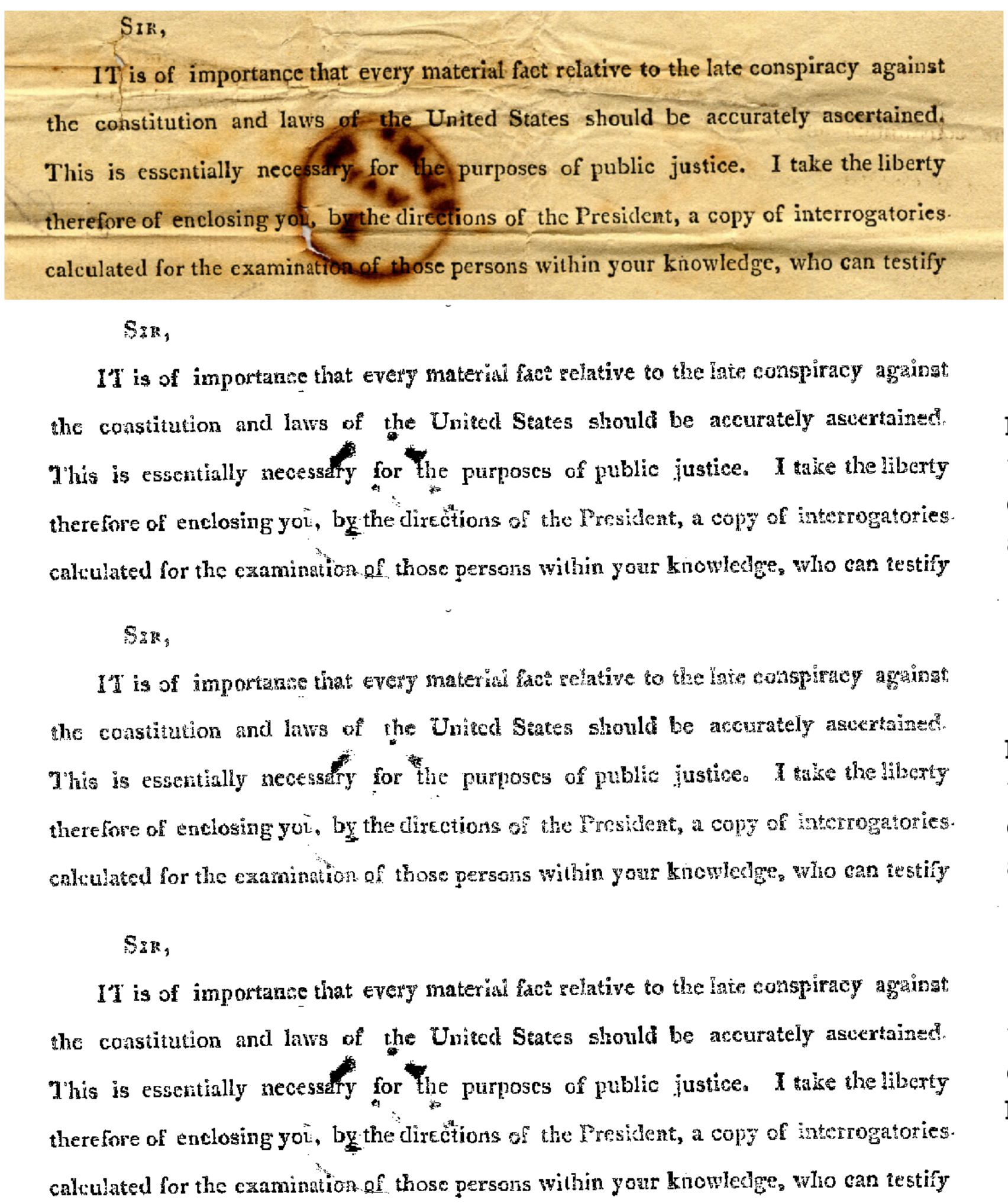

**N = 10; T0 = 0.2**
**tau = 0.125;**
**cd = 0.2; cs = 0.9;**
**sigma = 0.3; rho = 0.4;**

**N = 10; T0 = 0.2**
**tau = 0.125;**
**cd = 0.15; cs = 0.9;**
**sigma = 0.3; rho = 0.4;**

**tau = 0.125;**
**cd = 0.15; cs = 0.85;**
**N = 10; T0 = 0.2;**

Fig. 4. Results of PA using various parameter settings for degraded document image

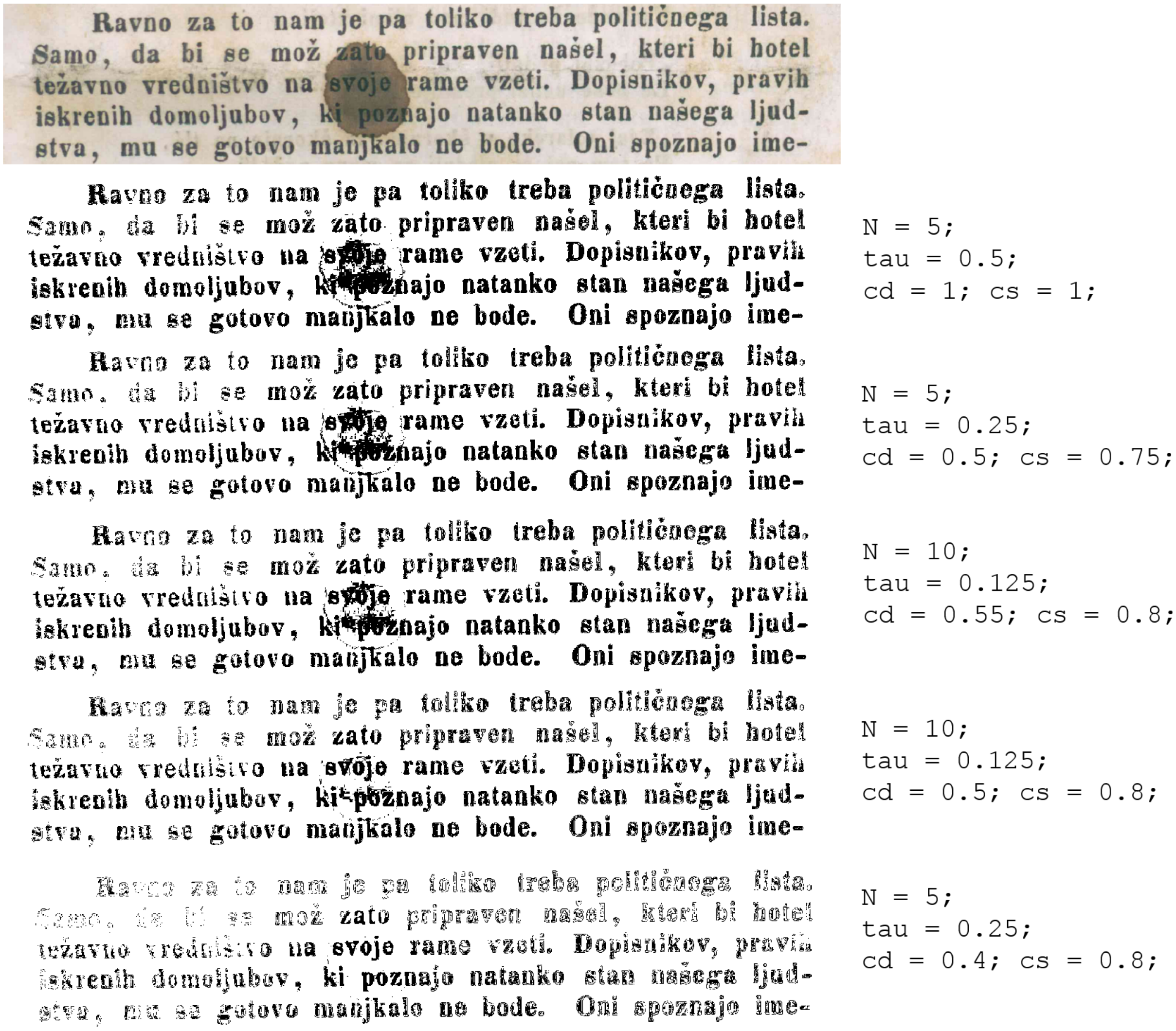

Fig. 5. Results of PA using various parameter settings for degraded document image with smudges

In the figures, cd = ce, which is the edge coefficient since there is no diffusion, thus cd = derivative or edge coefficient in this case. Thus, increasing the edge coefficient, removes noise and stains but also results in fading out of text with weak edges or where the stain edge profile overlaps with the text edge profile. Thus, the stains are eliminated but also at the cost of some text, which falls in the same intensity range of the stain. This issue was addressed somewhat but requires simultaneous stain removal with text enhancement, which the current model is ill-suited to perform.

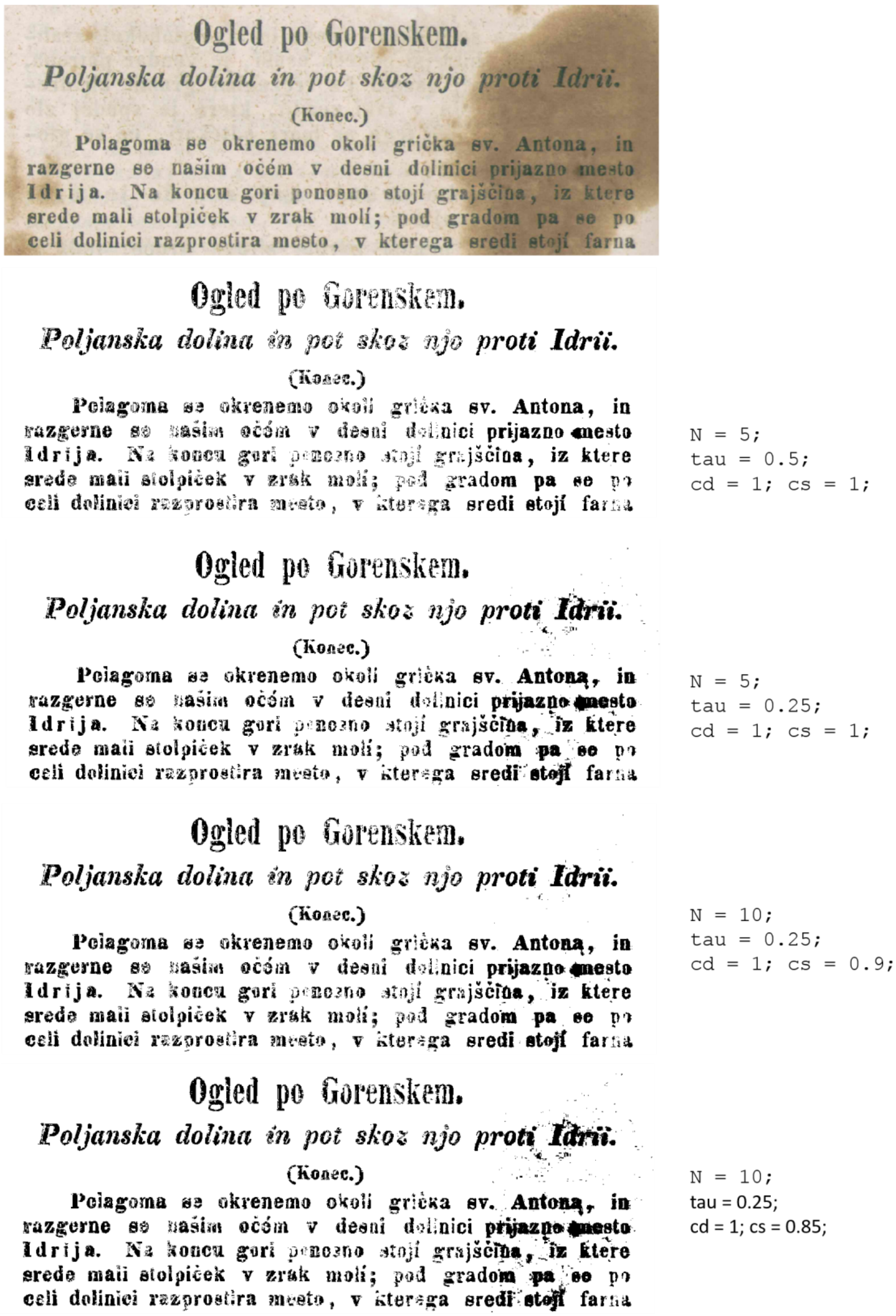

Fig. 6. Results of PA using various parameter settings for degraded document image with large oil stains

The numerical objective results using F-Measure (FM), pseudo F-measure (Fps), Peak Signal to Noise Ratio (PSNR), Distance Reciprocal Distortion (DRD) and Negative Rate Metric (NRM) [1] are presented in Table 1 from [1]. The larger the first three measures are, the better, while the reverse is the case of for the last two. The values suggest that the PA surpassed the PDE-based algorithms it was compared with.

Table 1 Performance comparison of PA with other PDE binarization algorithms

| **Metrics** / **Algorithms** | **FM (%)** | **$F_{PS}$ (%)** | **PSNR (dB)** | **DRD** | **NRM (%)** |
|---|---|---|---|---|---|
| WH | 80.37 | 85.17 | 16.25 | 7.95 | - |
| RMC | 75.29 | 77.62 | 15.15 | 18.33 | - |
| JM | 77.21 | 79.03 | 15.60 | 7.14 | - |
| WYH | 86.75 | 89.11 | 17.53 | 4.71 | - |
| FENG | 75.00 | 80.55 | 15.87 | 9.68 | - |
| GHZ | 85.78 | 88.44 | 17.60 | 5.75 | - |
| ZHG | 85.75 | 89.27 | 17.72 | 5.54 | - |
| GHW | 86.34 | 89.71 | 17.68 | 6.25 | - |
| PA (09216) | **88.00** | **90.30** | **18.11** | **3.99** | **0.07** |

## 5. Conclusion

The proposed PDE model is able to handle degraded document images better than the previous methods. However, it still has some shortcomings such as lost text in the attempt to remove stains. Some schemes were incorporated to address this issue but not always successful and requires constant tuning to achieve. Future work will focus on improving on the model to better handle such issues, without compromising quality and speed.

## References

[1] U. A. Nnolim, "Dynamic Selective Edge-Based Integer/Fractional Order Partial Differential Equation For Degraded Document Image Binarization," *International Journal of Image and Graphics*, vol. 21, no. 3, pp. 1-31, July 2021.

[2] U. A. Nnolim, "Improved integer/fractional order partial differential equation-based thresholding," *Optik - International Journal for Light and Electron Optics*, vol. 229, no. 2, pp. 1-10, March 2021.

[3] U. A. Nnolim, "Enhancement of degraded document images via augmented fourth order partial differential equation and Total Variation-based illumination estimation," *OPTIK*, vol. Accepted: In Press, 2021.

[4] J. Calvo-Zaragoza and A.-J. Gallego, "A selectional auto-encoder approach for document image binarization," *Pattern Recognition* , vol. 86 , p. 37–47, 2019.

[5] X. Huang, L. Li, R. Liu, C. Xu and M. Ye, "Binarization of degraded document images with global-local UNets," *Optik - International Journal for Light and Electron Optics*, vol. 203 , p. 164025, 2020.

[6] S. Kang, B. K. Iwana and S. Uchida, "Complex image processing with less data—Document image binarization by integrating multiple pre-trained U-Net modules," *Pattern Recognition* , vol. 109 , p. 107577, 2021.

[7] S. He and L. Schomaker, "DeepOtsu: Document enhancement and binarization using iterative deep learning," *Pattern Recognition* , vol. 91, p. 379–390, 2019.

[8] W. Xiong, J. Xu, Z. Xiong, J. Wang and M. Liu, "Degraded historical document image binarization using local features and support vector machine (SVM)," *Optik* , vol. 164 , p. 218–223, 2018.

[9] K. R. Ayyalasomayajula, F. Malmberg and A. Brun, "PDNet: Semantic segmentation integrated with a primal-dual network for document binarization," *Pattern Recognition Letters* , vol. 121 , p. 52–60, 2019.

[10] J. Zhao, C. Shi, F. Jia, Y. Wang and B. Xiao, "Document image binarization with cascaded generators of conditional generative adversarial networks," *Pattern Recognit.* , vol. 96 , 2019.

[11] J. Long, E. Shelhamer and T. Darrell, "Fully convolutional networks for semantic segmentation," in *CVPR*, 2015.

[12] J. Pastor-Pellicer, S. Espana-Boquera, F. Zamora-Martınez, M. Z. Afzal and M. Castro-Bleda, "Insights on the use of convolutional neural networks for document image binarization," in *13th International Work-Conference on Artificial Neural Networks*, 2015.

[13] O. Ronneberger, P. Fischer and T. Brox, "U-net: convolutional networks for biomedical image segmentation," in *Proceedings of the International conference on Medical Image Computing and Computer-Assisted Intervention*, 2015.

[14] J. Guo, C. He and X. Zhang, "Nonlinear edge-preserving diffusion with adaptive source for document images binarization," *Applied Mathematics and Computation,* vol. 351, pp. 8-22, 2019.

[15] S. S. Kumar, P. Rajendran, P. Prabaharan and K. Soman, "Text/Image region separation for document layout detection of old document images using non-linear diffusion and level set," in *6th International Conference on Advances in Computing and Communications (ICACC)*, 2016.

[16] F. Z. A. Bella, M. E. Rhabi, A. Hakim and A. Laghrib, "Reduction of the non-uniform illumination using nonlocal variational models for document image analysis," *J. Frankl. Inst.-Eng. Appl. Math*, vol. 355 , pp. 8225-8244, 2018 .

[17] K. Guemri and F. Drira, "Adaptative shock filter for image characters enhancement and denoising," in *6th International Conference of Soft Computing and Pattern Recognition (SoCPaR)*, 2014.

[18] Y. Wang, Q. Yuan and C. He, "Indirect diffusion-based level set evolution for image segmentation," *Appl. Math. Model.,* vol. 69 , p. 714–722, 2019.

[19] Y. Wang and C. He, "Binarization method based on evolution equations for document images produced by cameras," *Journal of Electronic Imaging,* vol. 21, no. 2, p. 023030 , 2012.

[20] B. A. Jacobs and E. Momoniat, "A novel approach to text binarization via a diffusion-based model," *Applied Mathematics and Computation*, vol. 225, pp. 446-460, 2013.

[21] B. A. Jacobs and E. Momoniat, "A locally adaptive, diffusion based text binarization technique," *Applied Mathematics and Computation*, vol. 269, pp. 464-472, 2015.

[22] D. Rivest-Hénault, R. Moghaddam and M. Cheriet, "A local linear level set method for the binarization of degraded historical document images," *Int. J. Doc. Anal. Recognit.* , vol. 15 , no. 2, p. 101–124, 2012.

[23] B. Chen, S. Huang, Z. Liang, W. Chen and B. Pan, "A fractional order derivative based active contour model for inhomogeneous image segmentation," *Appl. Math. Model.* , vol. 65, p. 120–136 , 2019.

[24] C. Huang and L. Zeng, "Level set evolution model for image segmentation based on variable exponent p-laplace equation," *Appl. Math. Model*, vol. 40 , no. 17–18, p. 7739–7750, 2016.

[25] J. Guo, C. He and Y. Wang, "Fourth order indirect diffusion coupled with shock filter and source for text binarization," *Signal Processing*, vol. 171 , no. 107478, pp. 1-13, January 2020.

[26] J. Guo and C. He, "Adaptive shock-diffusion model for restoration of degraded document images," *Applied Mathematical Modelling*, vol. 79, pp. 555-565, 2020.

[27] X. Zhang, C. He and J. Guo, "Selective diffusion involving reaction for binarization of bleed-through document images," *Applied Mathematical Modelling*, vol. 81, pp. 844-854, January 20 2020.

[28] S. Feng, "A novel variational model for noise robust document image binarization," *Neurocomputing*, vol. 325, pp. 288-302, October 6 2019.

[29] Z. Du and C. He, "Nonlinear diffusion equation with selective source for binarization of degraded document images," *Applied Mathematical Modelling* , vol. 99 , p. 243–259, July 2 2021.